\documentclass[11pt,a4paper]{article}
\usepackage[hyperref]{naaclhlt2019}
\usepackage{times}
\usepackage{latexsym}
\usepackage{amssymb}
\usepackage{mathtools}
\usepackage{url}
\usepackage{graphicx}
\usepackage{caption}
\usepackage{subcaption}
\usepackage{verbatim}
\usepackage[linesnumbered,lined,commentsnumbered]{algorithm2e}
\usepackage{tcolorbox}
\usepackage{color}
\usepackage{soul}
\usepackage[inline]{enumitem}
\usepackage{stmaryrd}
\usepackage{arydshln}
\usepackage{upgreek}

\usepackage{enumitem}

\aclfinalcopy 
\def\aclpaperid{***} 

\newcommand{\w}[1]{\textit{#1}}

\newcommand{\ra}[1]{$\xrightarrow{#1}$}
\newenvironment{termsetenv}{$\{$}{$\}$}
\newcommand{\termset}[1]{\begin{termsetenv}\term{#1}\end{termsetenv}}
\newcommand{\term}[1]{\textit{#1}}

\newcommand{\thrf}[2]{\ensuremath{\langle #1, #2\rangle}}

\newcommand{\mitof}{M^{\mathcal{V}\rightarrow\mathcal{F}}}
\newcommand{\mftoi}{M^{\mathcal{F}\rightarrow\mathcal{V}}}

\newcommand{\cbc}{CBC}
\newcommand{\cbcnn}{CBCNN}

\usepackage{xcolor}
\newcommand{\ignore}[1]{}

\title{ \vspace*{-0.5in}
{{\small \hfill NAACL'19}\\
\vspace*{.25in}} Text Classification with Few Examples using Controlled Generalization}

\author{Abhijit Mahabal \\
  Google, Pinterest \\
  {\tt amahabal@} \\
  {\tt gmail.com}\\\And
  Jason Baldridge \\
  Google\\
  {\tt jridge@} \\
  {\tt google.com}\\\And
  Burcu Karagol Ayan \\
  Google \\
  {\tt burcuka@}\\
  {\tt google.com}\\\And
  Vincent Perot \\
  Google \\
  {\tt vperot@} \\
  {\tt google.com}\\\And
  Dan Roth \\
  UPenn \\
  {\tt danroth@}\\
  {\tt seas.upenn.edu}}

\date{}

\begin{document}
\maketitle
\begin{abstract}

Training data for text classification is often limited in practice, especially for applications with many output classes or involving many related classification problems. This means classifiers must generalize from limited evidence, but the manner and extent of generalization is task dependent. Current practice primarily relies on pre-trained word embeddings to map words unseen in training to similar seen ones. Unfortunately, this squishes many components of meaning into highly restricted capacity. Our alternative begins with sparse pre-trained representations derived from unlabeled parsed corpora; based on the available training data, we select features that offers the relevant generalizations. This produces task-specific semantic vectors; here, we show that a feed-forward network over these vectors is especially effective in low-data scenarios, compared to existing state-of-the-art methods. By further pairing this network with a convolutional neural network, we keep this edge in low data scenarios and remain competitive when using full training sets.

\end{abstract}

\section{Introduction}

Modern neural networks are highly effective for text classification, with convolutional neural networks (CNNs) as the de facto standard for classifiers that represent both hierarchical and ordering information implicitly in a deep network \cite{kim:2014:EMNLP2014}. Deep models pre-trained on language model objectives and fine-tuned to available training data have recently smashed benchmark scores on a wide range of text classification problems \cite{peters-EtAl:2018:N18-1,howard-ruder:2018:Long,devlin2018bert}.

Despite the strong performance of these approaches for large text classification datasets, challenges still arise with small datasets with few, possibly imbalanced, training examples per class. Labels can be obtained cheaply from crowd workers for some languages, but there are a nearly unlimited number of bespoke, challenging text classification problems that crop up in practical settings \cite{yu-EtAl:2018:N18-1}. 
Obtaining representative labeled examples for classification problems with many labels, like taxonomies, is especially challenging. 

\textit{Text classification} is a broad but useful term and covers classification based on topic, on sentiment, and even social status. As Systemic Functional Linguists such as \newcite{halliday1985introduction} point out, language carries many kinds of meanings. For example, words such as \textit{ambrosial} and \textit{delish} inform us not just of the domain of the text (\textit{food}) and sentiment, but perhaps also of the age of the speaker. Text classification problems differ on the dimensions they distinguish along and thus in the words that help in identifying the class.

As \newcite{sachan-zaheer-salakhutdinov:2018:C18-1} show, classifiers mostly focus on sub-lexicons;  they memorize patterns instead of extending more general knowledge about language to a particular task. When there is low lexical overlap between training and test data, accuracy drops as much as 23.7\%. When training data is limited, most meaning-carrying terms are never seen in training, and the sub-lexicons correspondingly poorer. Classifiers must generalize from available training data, possibly exploiting external knowledge, including representations derived from raw texts. For small training sizes, this requires moving beyond sub-lexicons.

\begin{table*}[t]
	{\small\begin{tabular}{clcl}
			1.1&Kampuchea says \textbf{rice} crop in 1986 increased $\ldots$&2.1&\textbf{Gamma ray} Bursters. What are they?\\
			1.2&U.S. \textbf{sugar} policy may self-destruct $\ldots$&2.2&Life on \textbf{Mars}\\
			1.3&EC denies \textbf{maize} exports reserved for the U.S.$\ldots$&2.3&Single launch \textbf{space stations}\\
			1.4&U.S. \textbf{corn}, \textbf{sorghum} payments 50-50 cash/certs$\ldots$&2.4&\textbf{Astronauts}---what does weighlessness feel like?\\
			1.5&Canada \textbf{corn} decision unjustified$\ldots$&2.5&\textbf{Satellite} around \textbf{Pluto} mission?\\
	\end{tabular}}
	\caption{\textbf{Left}: examples from the Reuters \w{Grains} class, showing semantic type cohesion (kinds of crops). \textbf{Right}: post headers from the \w{sci.space} newsgroup in 20 Newsgroups, showing topical cohesion (astronomical terms). \textbf{Bolded terms} are to draw the reader's attention to parallels among examples.}\label{table:examples}
\end{table*}

Existing strategies for low data scenarios include treating labels as informative \cite{SongRo14,CRRS08} and using label-specific lexicons \cite{eisenstein2017unsupervised}, but neither is competitive when labeled data is plentiful. Instead, we seek classifiers that adapt to both low and high data scenarios.

People exploit parallelism among examples for generalization \cite{hofstadter2001analogy,hofstadter2013surfaces}. Consider Table \ref{table:examples}, which displays five examples from a single class for two tasks. Bolded terms for each task are clearly related, and to a person, suggest abstractions that help relate other terms to the task. This helps with disambiguation: that the word \w{Pluto} is the planet and not Disney's character is inferred not just by within-example evidence (e.g. \w{mission}) but also by cross-example presence of \w{Mars} and \w{astronauts}.

Cross-example analysis also reveals the amount of generalization warranted. For a word associated with a label, word embeddings give us neighbors, which often are associated with that label. What they do not tell us is the extent this associated-with-same-label phenomenon holds; that depends on the granularity of the classes. Cross-example analysis is required to determine how neighbors at various distances are distributed among labels in the training data. This should allow us to include \w{barley} and \w{peaches} as evidence for a class like \w{Agriculture} but only \w{barley} for \w{Grains}.

Most existing systems ignore cross-example parallelism and thus miss out on a strong classification signal. We introduce a flexible method for controlled generalization that selects syntacto-semantic features from sparse representations constructed by Category Builder \cite{mahabal2018robust}. Starting with sparse representations of words and their contexts, a tuning algorithm selects features with the relevant kinds and appropriate amounts of generalization, making use of parallelism among examples. This produces task-specific dense embeddings for new texts that can be easily incorporated into classifiers. 

Our simplest model, \cbc\ (Category Builder Classifier), is a feed-forward network that uses only CB embeddings to represent a document. For small amounts of training data, this simple model dramatically outperforms both CNNs and BERT \cite{devlin2018bert}. 
When more data is available, both CNNs and BERT exploit their greater capacity and broad pre-training to beat \cbc. We thus create \cbcnn, a simple combination of \cbc\ and the CNN that concatenates their pre-prediction layers and adds an additional layer. By training this model with a scheduled block dropout \cite{zhang-EtAl:2018:EMNLP1} that gradually introduces the \cbc\ sub-network, we obtain the benefits of \cbc\ in low data scenarios while obtaining parity with CNNs when plentiful data is available. BERT still dominates when all data is available, suggesting that further combinations or ensembles are likely to improve matters further.

\begin{table}
	\begin{tabular}{l|c|c|c}
		dataset&$k$&train/test/dev&size range\\\hline
		20NG&20&15076/1885/1885&513/810\\
		reuters&8& 6888/862/864 & 128/3128\\
		spam&2&3344/1115/1115&436/2908\\
		attack&2&10000/2000/2000&1126/8874\\
	\end{tabular}
	\caption{Data sizes, and the disparity between the smallest and the largest class in training data. The $k$ column indicates the number of classes in the task.}\label{table:data}
\end{table}

\section{Evaluation Strategy}

Our primary goal is to study classifier performance with limited data. To that end, we obtain learning curves on four standard text classification datasets (Table \ref{table:data}) based on evaluating predictions on the full test sets.  At each sample size, we produce multiple samples and run several text classification methods multiple times, measuring the following: 

\begin{itemize}
	\item \textbf{Macro-F1 score}. Macro-F1 measures support for all classes better than accuracy, especially with imbalanced class distributions.
	\item \textbf{Recall for the rarest class}. Many measures like F1 and accuracy often mask performance on infrequent but high impact classes, such as detecting toxicity \cite{waseem-hovy:2016:N16-2}) 
	\item \textbf{Degenerate solutions}. Complex classifiers with millions of parameters sometimes produce degenerate classifiers when provided very few training examples; as a result, they can skip some output classes entirely. 
\end{itemize}

The datasets we chose for evaluation, while all multi-class, form a diverse set in terms of the number of classes and kinds of cohesion among examples in a single class. The former clearly affects training data needs, while the latter informs appropriate generalization.

\begin{itemize}
	\item\textbf{20 Newsgroups} 20Newsgroups (20NG) contains documents from 20 different newsgroups with about 1000 messages from each. We randomly split the documents into an 80-10-10 train-dev-test split. The classes are evenly balanced.
	\item\textbf{Reuters R8.} The Reuters21578 dataset contains Reuters Newswire articles. Following several authors \cite[for example]{pinto2007relative,zhao2018topic}, we use only the eight most frequent labels. We begin with a given 80/10/10 split. Given that we focused on single-label classification, we removed items associated with two or more of the top eight labels (about 3\% of examples). Classes are highly imbalanced. Of the 6888 training examples, 3128 are labeled \w{earn}, while only 228 examples are of class \w{interest} and only 128 are \w{ship}.
	\item\textbf{Wiki Comments Personal Attack.} The Wikipedia Detox project collected over 100k discussion comments from English Wikipedia and annotated them for presence of personal attack \cite{wulczyn2017ex}. We randomly select 10k, 2k, and 2k items as train/dev/test. 11\% are attacks.
	\item \textbf{Spam} The SMS Spam Collection v.1 has SMS labeled messages that were collected for mobile phone spam research \cite{hidalgo2012validity}. Each of the 5574 messages is labeled as \w{spam} or \w{ham}.
\end{itemize}

\begin{figure}[t]
	\centering\includegraphics[width=0.45\textwidth]{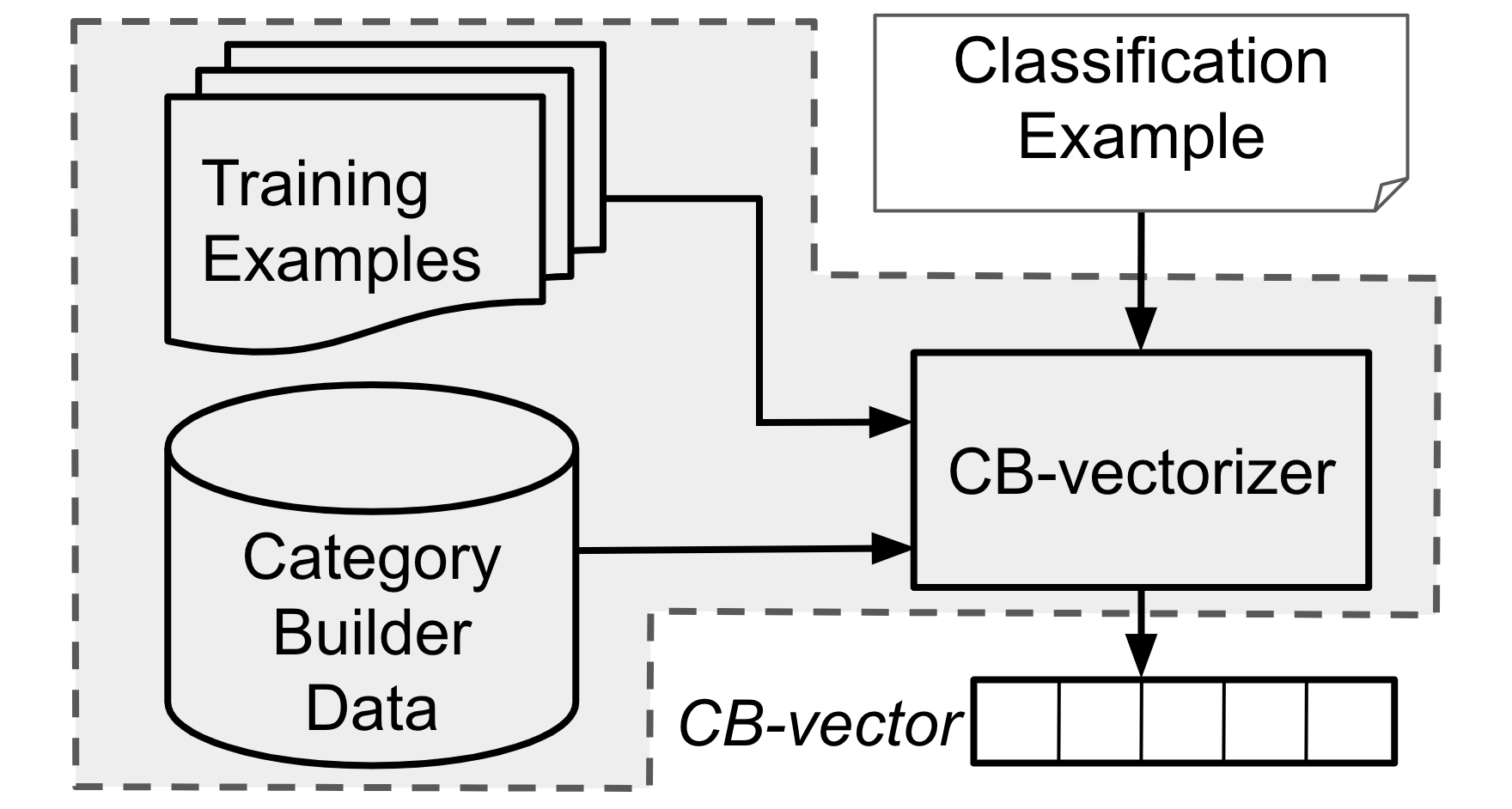}
	\caption{\textbf{Shaded Region:} We use Category Builder data \cite{mahabal2018robust} and identify generalizing features in training data, producing a \w{vectorizer}. This is done once. \textbf{Unshaded:} Given a document, the vectorizer produces a dense vector usable in deep networks.}\label{fig:flow}
\end{figure}

\section{Identifying Generalizing Features}\label{sec:generalizing}

\begin{table}[t]
	\centering
	\begin{tabular}{rp{1.9in}}
		\textbf{Feature}&\textbf{Prototypical Supports}  \\\hline
		\w{allergen as X}& \textit{pollen, dander, dust mites, soy, perfumes, milk, smoke, mildew}\\
		\w{liter of X}& \textit{water, petrol, milk, fluid, beer}\\
		\w{serve with X} & \textit{rice, sauce, salad, fries, milk} \\\hdashline
		\w{flour mixture} & \textit{butter mixture, rubber spatula, dredged, creamed, medium speed, sifted, milk}\\
		\w{replacer} & \textit{colostrum, calves, whole milk, inulin, pasteurized, weaning}\\
	\end{tabular}
	\caption{A few features (among hundreds) evoked by \w{milk}, with top n-grams in their support. \textbf{Above dashed line ($F_S$)} fit in tidy categories (here, \w{allergen}, \w{fluid}, and \w{food} are rough glosses). \textbf{Below dashed line ($F_C$)} are not describable by simple labels---the evoking terms have different parts of speech and instead display \w{situational coherence}, e.g. association with the process of mixing flour or with animal husbandry (a \w{replacer} is milk formula for calves).}
	\label{tab:milk}
\end{table}

In this section, we explicate the source of features, discuss the properties relevant to generalization by focusing on one feature in isolation, and present the overall feature selection method. The overview in Figure \ref{fig:flow} displays the order of operations: identify \w{generalizing features} based on the training data (done once), and for each document to be classified, convert it to a vector, where each entry corresponds to a generalizing feature.

\subsection{Category Builder}

Our source of generalizing features is Category Builder (CB) \cite{mahabal2018robust}, which constructs a sparse vector space derived from parsed corpora \cite{erk:2012}. CB constructs features for \textit{n-grams} (not just unigrams) that are the union of syntactic context features $F_S$ and co-occurrence features $F_C$. Consider \w{milk}: an $F_S$ feature is \w{gallon{\ra{prep}}of{\ra{pobj}}X} and $F_C$ features include \w{goat}, \w{cow}, \w{drink}, \w{spill}, etc. Table \ref{tab:milk} provides other examples of features evoked by \w{milk}, along with other n-grams which evoke them. For present purposes, we can treat CB as a matrix with n-grams as rows and features in $F_C$ and $F_S$ as columns. The entries of CB are weights that give the association strength between an n-gram and a feature; these weights are an asymmetric variant of pointwise-mutual information \cite{mahabal2018robust}.

\subsection{Properties of Generalizing Features}

Which features generalize well depends on the granularity of classes in a task. Useful features for generalization strike a balance between \w{breadth} and \w{specificity}. A feature that is evoked by many words provides generalization potential because the feature's overall support is likely to be distributed across both the training data and test data. However, this risks over-generalization, so a feature should also be sufficiently specific to be a precise indicator of a particular class.

A key aspect of choosing good features based on a limited training set is to resolve referential ambiguity \cite{quine1960word,wittgenstein1953philosophical} to the extent supported by the observed uses of the words.
To illustrate, consider the \w{grains} class in the Reuters Newswire dataset. The word \w{wheat} can evoke the features at different levels of the taxonimical hierarchy: \w{triticum} (the wheat genus), \w{poaceae} (grass family), \w{spermatophyta} (seeded plants), \w{plantae} (plant kingdom), and \w{living thing}. The first among these has low breadth and is evoked only by \w{wheat}. The second is far more useful: specific and yet with a large support, including \w{maize} and \w{sorghum}. The final feature is too broad. In general, the most useful features for generalization are the intermediate features, also known as Basic Level Categories \cite{rosch1976basic}.

Another important aspect of generalization comprises the \textit{facets} of meaning. For example, the word \w{milk} has facets relating it to other liquids (e.g., \w{oil}, \w{kerosene}), foods (\w{cheese}, \w{pasta}), white things (\w{ivory}), animal products (\w{honey}, \w{eggs}), and allergens (\w{pollen}, \w{ragweed}). Along these axes, generalization can be more or less conservative; e.g., both \w{cheese} and \w{tears of a phoenix} are animal products, but the former is semantically closer to \w{milk}.  Looking back at Table \ref{tab:milk}, the utility of individual features evoked by \w{milk} for tasks involving related topics varies; e.g., does the classification problem pertain to \w{food} or \w{animal husbandary}?

\subsection{Focus on a single feature}\label{sec:overview}

A single \w{generalizing feature} is associated with many n-grams, each of which evokes it (with different strengths). Table \ref{tab:support} displays n-grams that evoke the feature \textit{co-occurrence with \textbf{\w{Saturn V}}}, as discovered by unsupervised analysis of a large corpus of web pages. The table further displays the interaction of this unsupervised feature  with supervised data, specifically, with the label \w{sci.space} in 20NG, when using a size 320 training sample that contain only 18 \w{sci.space} documents. Counts for some evoking terms are shown within and outside this class, for both training and test data.

\textbf{Notation.} We introduce some notation and explicate with Table \ref{tab:support}.  We have a labelled collection of training documents $T$.  $T_l$ is the training examples with label $l$. The \w{positive support set} $\Psi_l(f,t)$ is the set of n-grams in $T_l$ evoking feature $f$ with weight greater than $t$, here, \termset{apollo, rocket, \ldots, shuttle} for $t{=}2.3$. The \w{positive support size} $\Lambda_l(f,2.3){=}|\Psi_l(f,2.3)|{=}5$ and the \w{positive support weight} $\lambda_l(f,2.3)$ is the sum of counts of supports of $f$ in $l$ with weight greater than $2.3$, here $1{+}3{+}2{+}1{+}2{=}9$. Analogously the \w{negative support weight} $\lambda_{\overline{l}}(f,2.3)$ is the sum of counts from outside $T_l$; here, $1{+}1{=}2$ since \termset{apollo, rocket} were seen outside \w{sci.space} once each.

\begin{table}[t]
	\centering
	\begin{tabular}{c|c|cc|cc}
		&&\multicolumn{2}{c|}{Training}&\multicolumn{2}{c}{Testing}\\
		\textbf{n-gram}&wt&$C$&$\overline{C}$&$C$&$\overline{C}$\\\hline
		apollo&8.93&1&1&5&1  \\
		\textbf{launch pad}&8.52&0&0&1&0\\ 
		rocket&7.32&3&1&8&0\\
		rockets&7.27&2&0&4&1\\
		liftoff&6.92&1&0&1&0\\
		\textbf{space shuttle}&6.27&0&0&4&0\\
		\textbf{space station}&6.19&0&0&4&3\\
		\textbf{payload}&4.23&0&0&5&0\\
		shuttle&2.57&2&0&15&3\\\hdashline
		kennedy&2.30&1&0&1&4\\
		capacity&1.95&0&1&0&4\\
	\end{tabular}
	\caption{Some evoking n-grams associated with the CB feature \textit{co-occurrence with \textbf{\w{Saturn V}}} and pivoting on the class \w{sci.space}. Counts for n-grams in training (sample size 320) and test data are shown, within \w{sci.space} ($C$) and outside ($\overline{C}$). Bolded n-grams are not seen in training but occur in test, providing generalization. The dashed line represents a threshold; higher scoring n-grams are more cohesive, and thresholding can make a feature cleaner by decreasing semantic drift.}
	\label{tab:support}
\end{table}

What makes this feature (\textit{words that have co-occurred with Saturn V}) well suited for \w{sci.space} is that many evoking words here are associated with the label \w{sci.space}. What confirms the benefit is the limited amount of negative support. Crucially, the bolded terms do not occur in the training data, but do occur in the test data. (We stress that we include these counts here only for this example; our methods do not access the test data for feature selection in our experiments.)

That said, we must limit potential noise from such features, so we seek \textbf{thresholded features} $\thrf{f}{t}$, as suggested by the dashed line in Table \ref{tab:support}. Items below this line are prevented from evoking $f$. We choose the highest threshold such that dropped negative support exceeds dropped positive support. This is determined simply by going through all the supports of a feature, sorted by ascending weight, and checking the positive and negative support of all features with smaller versus greater weight given the class. The weight of the feature at this cusp is used as the threshold of the feature for this particular class. This $\thrf{f}{t}$ pair then forms one element of the CB-vector used as a feature for classification. 

Given the labeled subsets of $T$ and this feature thresholding algorithm, we produce a vectorizer that embeds documents. The values of a document's embedding \w{are not} directly associated with any class. Such association happens during training. Although \w{sci.space} accounts for just 6\% of the documents, 75\% of documents that contain an n-gram evoking the \textit{Saturn V} feature are in that class. A classifier trained with such an embedding should learn to associate this feature with that class, and an unseen document containing the unseen-in-training term \w{space shuttle} stands a good chance to be classified as \w{sci.space}.

The feature displayed in Table \ref{tab:support} is useful for the 20NG problem because it contains a class related to space travel. This feature has no utility in spam classification or in sentiment classification, since, for those problems, seeing \w{rocket} in one class does not make it more likely that a document containing \w{space station} belongs to that same class. This example illustrates why a generalization strategy must incorporate both what we can learn from unsupervised data as well as (limited) labeled training data.

\subsection{Overall feature selection}

We now describe how we use the training data $T$ to produce a set of features-and-threshold pairs; each chosen feature-with-threshold $\thrf{f}{t}$ will be one component in the CB-vectors provide to classifiers. Calculation of features for a single class is a three step process: \begin{enumerate*}[label=(\roman*)]
	\item for each feature $f$, choose a threshold $t$ (as discussed above)
	\item score the resultant $\thrf{f}{t}$
	\item filter useless or redundant $\thrf{f}{t}$.
\end{enumerate*}

Given a label $l$ and a feature $f$, we implicitly produce a table of supporting n-grams and their distribution within and outside $l$ (e.g. as in Table \ref{tab:support}). This involves computing the precision of a feature at a given threshold value, comparing it to the class probability and deciding whether to keep it.

Recall the positive support $\lambda_l(f,t)$ and negative support $\lambda_{\overline{f}}(f,t)$ defined previously.  The \textbf{precision} of $f$ at threshold $t$ is $\mu_l(f,t)=\frac{\lambda_l(f,t)}{\lambda_l(f,t)+\lambda_{\overline{l}}(f,t)}$,  (this is $\frac{9}{11}$ in the example of Table \ref{tab:support}, with $t{=}2.3$). However, since we are often dealing with low counts, we smooth the precision toward the empirical class probability of $l$, $p(l)=\frac{|T_l|}{|T_l|+|T_{\overline{l}}|}$.

\[ \tilde{\mu}_l(f,t)=\frac{\lambda_l(f,t)+p(l)\alpha}{\lambda_l(f,t)+\lambda_{\overline{l}}(f,t)+\alpha} \]

The score $S_l(f,t)$ is reduction in error rate of the smoothed precision relative to the base rate:

\[S_l(f,t)=\frac{\tilde{\mu}_l(f,t)-p(l)}{1-p(l)} \]

\noindent We retain a thresholded feature if it is generalizing ($\Lambda_l(\thrf{f}{t}) > 1$), has better-than-chance precision (we use $S_l(\thrf{f}{t}) > 0.01$), and is not redundant (i.e., its positive support has one or more terms not present in positive supports of higher scoring features).

\subsection{Creating the CB-vector}

Each vector dimension corresponds to some $\thrf{f}{t}$. The evocation level of $f$ is the sum of its evocation for the n-grams in the document $d$, $e_d(f) = \sum_{w\in d}CB(w, f)$. The vector entry is $\frac{e_d(f)}{t}$ when $e_d(f) >= t$, and is clipped to $0$ otherwise.

\section{Models}
\label{sec:models}

As benchmarks, we use a standard CNN with pretrained embeddings \cite{kim:2014:EMNLP2014} and BERT \cite{devlin2018bert}.\footnote{\scriptsize{\url{ https://github.com/google-research/bert}}} For CNN, we used 300 filters each of sizes 2, 3, 4, 5, and 6, fed to a hidden layer of 200 nodes after max pooling. Pretrained vectors provided by Google were used.\footnote{\scriptsize{\url{https://code.google.com/archive/p/word2vec}}} For BERT, we used the \texttt{run\_classifier} script from GitHub and used the BERT-large-uncased model.

We use the pre-computed vocab-to-context association matrix provided as part of the open source Categorial Builder repository.\footnote{\scriptsize \url{https://github.com/google/categorybuilder}} This contains 194,051 co-occurrence features ($F_C$) and 954,276 syntactic features ($F_S$).

\textbf{\cbc\ model}. The \w{CB-vector} containing the derived features from the training dataset and Category Builder can be exploited in various ways with existing techniques. The simplest of these is to use a feed-forward network over the \w{CB-vector}. This model does not encode the tokens or any word order information---information which is highly informative in many classification tasks. 

\textbf{\cbcnn\ model}. Inspired by the combination of standard features and deep networks in Wide-and-Deep models \cite{Cheng:2016:WDL:2988450.2988454}, we pair the \cbc\ model with a standard CNN, concatenating their pre-prediction layers, and add an additional layer before the softmax prediction. In early experiments, this combined model performed worse than the CNN on larger data sizes, as the network above the CB-vector effectively stole useful signal from the CNN. To ensure that the more complex CNN side of the network had a chance to train, we employed a block dropout strategy \cite{zhang-EtAl:2018:EMNLP1} with a schedule. During training, with some probability, all weights in the CB-vector are set to 0.5. The probability of hiding decreases from 1 to 0 using a parameterized hyperbolic tangent function $p_k{=}\frac{2}{e^{Cx} + 1}$. Lower values of $C$ lead to slower convergence to zero. The effect is that the \cbc\ sub-network is introduced gradually, allowing the CNN to train while eventually taking advantage of the additional information.

The natural strategy of replacing with 0s (instead of 0.5 as above) was tried and also works, but less well, since the network has no way to distinguish between genuine absence of feature and hiding. In CB-vector, non-zero values are at least 1, and thus 0.5 does not suffer from this problem.

\section{Experiments}

Our primary goal is to improve generalization for low-data scenarios, but we also want our methods to remain competitive on full data.

\subsection{Experimental setup}

We compare different models across learning curves of increasing the training set sizes. We use training data sizes of $40, 80, \ldots, 5120$ as well as the entire available training data. For each training size, we produce three independent samples by uniformly sampling the training data and training each model three times at each size. The final value reported is the average of all nine runs.  All models are implemented in Tensorflow. Batch sizes are between 5 and 64 depending on training size. Training stops after there is no macro-F1 improvement on development data for 1000 steps.

For evaluation, we focus primarily on macro-F1 and recall of the rarest class. The recall on the rarest class is especially important for imbalanced classification problems. For such problems, a model can obtain high accuracy by strongly preferring the majority class, but we seek models that effectively identify minority class labels. (This is especially important for active learning scenarios, where we expect the CB-vectors to help with in future.)

\begin{figure*}
	\centering\includegraphics[width=0.95\textwidth]{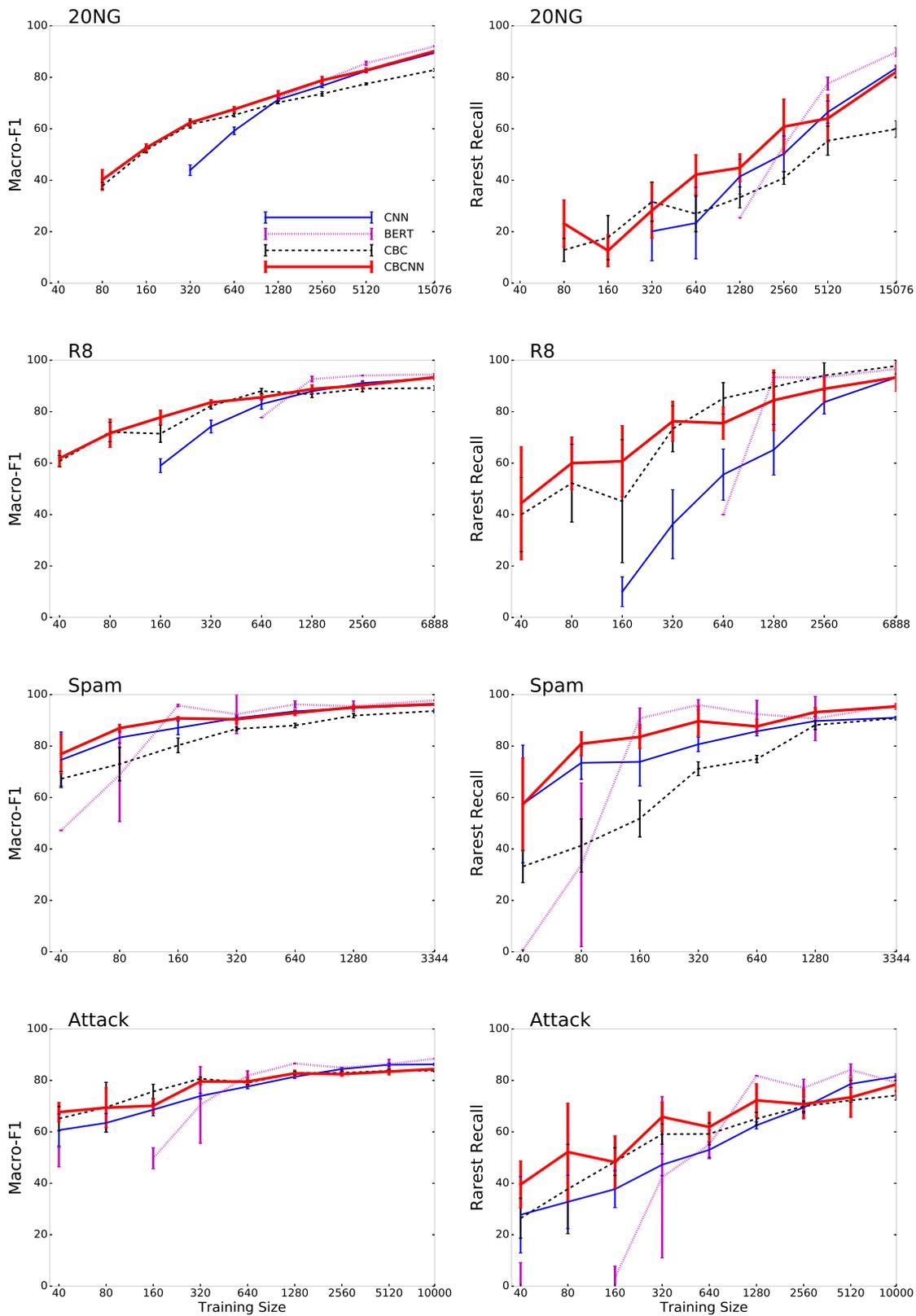}
	\caption{\textbf{Left:} F1 score by training size for 20NG, Reuters, SMS Spam, and Wiki-attack. Data is shown for non-degenerate models, and hence CNN and BERT start at higher sizes (see Table \ref{tab:min_size}). \textbf{Right:} Recall for the rarest class for the same models.}\label{fig:four}
\end{figure*}

\subsection{Results: low data scenarios}

Figure \ref{fig:four} shows learning curves giving macro-F1 scores and rarest class recall for all four datasets. When very limited training data is available, the simple \cbc\ model generally outperforms the CNN and BERT, except for the Spam dataset. The more powerful models eventually surpass \cbc; however, the \cbcnn\ model provides consistent strong performance at all dataset sizes by combining the generalization of \cbc\ with the general efficacy of CNNs. Importantly, \cbcnn\ provides massive error reductions with low data for 20NG and R8 (tasks with many labels).

Table \ref{tab:full}'s left half gives results for all models when using only 320 training examples. For 20NG, CNN's macro-F1 is just 43.9, whereas \cbc\ and \cbcnn\ achieve 61.7 and 62.4---the same as CNN performance with four times as much data. These models outperform CNN on R8 as well, reaching 83.7 vs CNN's 74.1, and also on the Wiki-attack dataset, achieving 80.6 vs CNNs 74.0. BERT fails to produce a solution for the two datasets with $>$2 labels, but does produce the best result for Spam---indicating an opportunity to more fully explore BERT's parameter settings for low data scenarios and to fruitfully combine \cbc\ with BERT.

\begin{table*}[t]
	\centering
	\begin{tabular}{c|c|cccc|cccc}
		&&\multicolumn{4}{c|}{320 Training Examples}&\multicolumn{4}{c}{Full Training Data}\\
		{Data}&k&{\cbc}&{CNN}&{\cbcnn}&{BERT} &{\cbc}&{CNN}&{\cbcnn}&{BERT} \\\hline
		20NG&20&61.7&43.9&\textbf{62.4}&---& 82.9&89.5&90.2&{\bf 92.0}\\
		R8&8&82.3&74.2&\textbf{83.6}&---&89.2&93.1&93.4&{\bf 94.4}\\
		Spam&2&80.3&87.1&90.8&\textbf{95.7}& 93.7 & 96.1 & 96.3 & {\bf 97.8}\\
		Attack&2&\textbf{80.7}&74.0&79.6&70.5  &83.8&86.2&84.5&{\bf 88.2}\\
	\end{tabular}
	\caption{Macro-F1 scores on all data sets when using 320 training examples (left) and when using all available training data (right). $k$ is the number of classes. The \cbcnn\ model provides the strongest overall performance across all data sizes. (Note that BERT produces degenerate solutions for the $>$2 class problems with 320 examples.)}
	\label{tab:full}
\end{table*}

\begin{table}
	\centering
	\begin{tabular}{c|c|cccc}
		Model&k&\cbc&CNN&\cbcnn&BERT\\\hline
		20NG&20&80&320&80& 1280 \\
		R8&8&40&160&40&640 \\
		Spam&2&40&40&40&40\\
		Attack&2&40&40&40&40\\
	\end{tabular}
	\caption{Minimum training size at which a non-degenerate model was produced in any of 9 runs. With more classes, more data is needed by CNN and BERT to produce acceptable models. k is number of classes.}
	\label{tab:min_size}
\end{table}

Rarest class recall is generally much better with less data when exploiting CB-features. For example, with 320 training examples for R8, CNNs reach 36.2 whereas \cbcnn\ scores 76.2. Prediction quality with few training examples (especially getting good balance across all labels) also interacts with other strategies for dealing with limited resources, such as active learning. For example, \newcite{baldridge_osborne_2008} obtained stronger data utilization ratios with better base models and uncertainty sampling for Reuters text classification: better models pick better examples for annotation and thus require fewer new labeled data points to achieve a given level of performance.

Importantly, the \cbc\ and \cbcnn\ models take far less data to produce non-degenerate models (defined as a model which produces all output classes as predictions). CNN and BERT have a large number of parameters, and using these powerful tools with small training sets produces unstable results. Table \ref{tab:min_size} gives the minimum training set sizes at which each model produces at least one non-degenerate model. While it might be possible to ameliorate the instability of CNN and BERT with a wider parameter search and other strategies, nothing special needs to be done for \cbc. It is likely that an approach which adaptively selects \cbc\ or \cbcnn\ and BERT would obtain the strongest result across all training set sizes.

For each dataset, among the 100 best features chosen (for training size 640), the breakdown of domain features ($F_C$) versus type features ($F_S$) is revealing. As expected, domain features are more important in a topical task such as 20NG (71\% are $F_C$ features), while the opposite is true for Spam (19\%) and a toxicity dataset like Wiki Attack (23\%). Reuters shows a fairly even balance between the two types of features (41\%): it is useful for R8 to be topically coherent and also to hone in on fairly narrow groups of words that collectively cover a Basic Level Category.

\subsection{Results: full data scenarios}

Table \ref{tab:full} provides macro-F1 scores for all models when given all available training data. The \cbc\ model performs well, but its (intentional) ignorance of the actual tokens in a document takes a toll when more labeled documents are available. The CNN benchmark, which exploits both word order and the tokens themselves, is a strong performer. The \cbcnn\ model effectively keeps pace with the CNN---improving on 20NG and R8, though slipping on Wiki-Attack. BERT simply crushes all other models when there is sufficient training data, showing the impact of structured pre-training and consistent with performance across a wide range of tasks in \newcite{devlin2018bert}.

\section{Conclusion}

We demonstrate an effective method for exploiting syntactically derived features from large external corpora and selecting the most useful of those given a small labeled text classification corpus.  We show how to do this with the map provided by Category Builder n-grams to features, but other sources of well generalizing features have been exploited for text classification. These include topic models \cite{blei2003latent}, ontologies such as WordNet \cite{bloehdorn2004boosting} and Wikipedia Category structure \cite{gabrilovich2009wikipedia}. It may be possible to use these other sources exactly as we use CB. Some of these sources have been manually curated, which makes them high quality but limits the size and facets. We have not yet explored their use because CB features seem to cover many of these sources' strengths---for example, $F_C$ features are like topics, and $F_S$ features like nodes in ontologies. Nonetheless, a combination may add value.

Our focus is on data scarce scenarios. However, it would be ideal to derive utility at both the small and large labeled data sizes. This will likely require models that can generalize with contextual features while also exploiting implicit hierarchical organization and order of the texts, e.g. as done by CNNs and BERT. The \cbcnn\ model is one effective way to do this and we expect there could be similar benefits from combining \cbc\ with BERT. Furthermore, approaches like AutoML \cite{zoph:le:2017} would likely be effective for exploring the design space of network architectures for representing and exploiting the information inherent in both signals.

Finally, although we focus on multi-class problems here---each example belongs to a single class---the general approach of selecting features should work for multi-label problems. Our confidence in this (unevaluated) claim stems from the observation that we select features one class at a time, treating that class and its complement as a binary classification problem. 

\section*{Acknowledgements}
We would like to thank our anonymous reviewers and the Google AI Language team, especially Rahul Gupta, Tania Bedrax-Weiss and Emily Pitler, for the insightful comments that contributed to this paper.

\bibliography{naaclhlt2019}
\bibliographystyle{acl_natbib}

\appendix
\section{Appendix}
\label{sec:appendix}

\subsection*{Changes to the Category Builder Matrix}

Category Builder uses two matrices: one mapping items to features ($\mitof$), the other mapping features to items ($\mftoi$). Two are needed since the relationship is asymmetric: the feature \textit{X is a star sign} is more strongly associated with the term \textit{cancer} than vice versa, and the two matrices are thus not exact transposes of each other, although they almost are. For this current work, we just use one matrix, $\mitof$. For syntactic features $F_S$, we directly use the Category Builder rows. For $F_C$ features, however, Category Builder replaced the corresponding submatrix in $\mitof$ with an identity matrix as described in \cite{mahabal2018robust}. We obtain that part of the matrix by copying the corresponding rows from $(\mftoi)^T$.

This new matrix will be made available as part of the Category Builder project.

\end{document}